\documentclass[conference]{IEEEtran}
\usepackage{subcaption}
\usepackage[numbers]{natbib}
\usepackage{xcolor}
\usepackage{lmodern}
\usepackage[nocref]{alexmacros}
\usepackage{paralist}
\usepackage{graphicx}
\usepackage{todonotes}
\usepackage[bookmarks=true]{hyperref}
\usepackage{algorithm}
\usepackage[noend]{algpseudocode}

\graphicspath{{figures/}}
\hypersetup{pdfpagemode=UseNone}

\begin{document}
\title{Long-Lived Distributed Relative Localization of Robot Swarms}
\author{
\authorblockN{Alejandro Cornejo}
\authorblockA{School of Engineering and Applied Sciences\\
  Harvard University\\
  Cambridge MA 02138}
\and
\authorblockN{Radhika Nagpal}
\authorblockA{School of Engineering and Applied Sciences\\
  Harvard University\\
  Cambridge MA 02138}}

\maketitle

\begin{abstract}
This paper studies the problem of having mobile robots in a multi-robot 
system maintain an estimate of the relative position and relative 
orientation of near-by robots in the environment. This problem is 
studied in the context of large swarms of simple robots which are 
capable of measuring only the distance to near-by robots.

We present two distributed localization algorithms with different 
trade-offs between their computational complexity and their 
coordination requirements.
The first algorithm does not require the robots to coordinate their 
motion. It relies on a non-linear least squares based strategy to 
allow robots to compute the relative pose of near-by robots.
The second algorithm borrows tools from distributed computing theory 
to coordinate which robots must remain stationary and which robots are 
allowed to move. This coordination allows the robots to use standard 
trilateration techniques to compute the relative pose of near-by 
robots.
Both algorithms are analyzed theoretically and validated through 
simulations.
\end{abstract}


\section{Introduction}
Most tasks which can be performed effectively by
a group of robots require the robots to have some information about the 
relative positions and orientations of other nearby robots.
For example in flocking~\cite{flock} robots use the relative orientation 
of each of its neighbors to control their own heading, in formation 
control~\cite{shape} robots control their own position as a function of 
the relative position of their neighbors, and in mapping~\cite{explore} 
robots use the relative position and relative orientation of their 
neighbors to interpret the information collected by neighboring robots.
However, most of the existing work on localization addresses 
localization of a single robot, requires landmarks with known positions 
on the environment, or relies on complex and expensive sensors.
Many environments of interest prevent the use of landmarks, and complex 
and/or costly sensors are not available in swarm platforms, which are 
composed of large numbers of low-cost robots.

We study the problem of having each robot in a multi-robot system 
compute the relative pose (position and orientation) of close-by robots 
relying only on distance estimates to close-by robots.
The algorithms described in this paper are fully distributed, and the 
computations performed at each robot rely only on information available 
in its local neighborhood.
This problem is long-lived, since for any mobile robot, the set of 
close-by robots and their relative pose changes during the execution.
We consider a general problem formulation which does not require
explicit control over the motions performed by the robots. This allows 
composing solutions to this problem with motion-control algorithms to 
implement different higher-level behaviors.
Furthermore, we study this problem in a robot swarm setting, which 
imposes sensor and computational restrictions on the solutions.
The table below summarizes the communication and computational
complexity requirements of the two distributed algorithms proposed in 
this paper.

\begin{figure}[htpb]
  \centering
  \begin{tabular}{r|c|c}
    & Communication & Computation \\ \hline
    Algorithm 1 & $O(1)$ & $O(\varepsilon^{-2})$ \\
    Algorithm 2 & $O(\Delta)$ & $O(1)$
  \end{tabular}
  \caption{Communication and computational requirements of the 
    algorithms proposed in this paper. Communication costs are measured 
    per round, and computational costs are per round per robot 
    localized. $\Delta$ denotes the maximum degree of the graph, and 
    $\varepsilon$ represents the error.}
\end{figure}

In a typical swarm platform, the communication, computation and sensing 
capabilities of individual robots are fairly limited.
The communication limitations of the individual robots in a swarm 
platform rule out any strategy that requires collecting large amounts of 
data at hub locations, and yet, the simplicity of the individual robots 
demand some form of cooperation. Moreover, the computational constraints 
of individual robots exclude the possibility of storing and updating 
complex models of the world or other robots.

Therefore, to fully exploit the potential of a robot swarm platform, it 
is paramount to use decentralized strategies that allow individual 
robots to coordinate locally to complete global tasks.
This is akin to the behavior observed in swarms of insects, which 
collectively perform a number of complex tasks which are unsurmountable 
by a single individual, all while relying on fairly primitive forms of 
local communication. 

\subsection{Related work}

The problem of localization using distance sensors has received a lot of
attention, most of it focusing on landmark-based localization\footnote{
Landmark-based localization assumes the environment contains a set of 
landmarks with known positions, and to which the robots can measure 
either the distance or the angle. GPS is an example of this type of 
localization, where satellites on known orbits play the role of 
landmarks. It is also common for a subset of robots or sensors with 
known positions to be the landmarks.}.
Using only connectivity information to stationary landmarks with known 
positions~\cite{globecom07}, it is possible to approximate the position 
of mobile nodes.
When distance measurements to the landmarks are available, systems such 
as the Cricket Location-Support System (which uses ultrasound sensors) 
can localize mobile nodes within a predefined region, and it has been 
shown how to obtain finer grained position information using a similar 
setup~\cite{dynamic}.

The more general case of fixed stationary landmarks with unknown initial 
positions has also been considered in the 
literature~\cite{icra06,olson06}. The case where the set of robots that 
play the role of stationary landmarks changes through an execution has 
also been considered~\cite{beacon99}, but in contrast to the present 
work, it requires knowledge of the initial landmark positions and 
provides no explicit mechanism for coordinating which robots play the 
role of landmarks.

One of the few landmark-free localization methods is the robust 
quadrilaterals work~\cite{moore04}, which is based on rigidity theory.  
However, in contrast to the present work it is designed primarily for 
static sensor networks and does not recover the relative orientation of 
nodes.

\subsection{Road map.} Section~\ref{sec:model} describes the formal 
system model and problem formulation. Sections~\ref{sec:alg1} 
and~\ref{sec:alg2} present and analyze two different algorithms for the 
localization problem. Finally Section~\ref{sec:exp} evaluates the 
performance of these algorithms through simulations.

\section{System Model}
\label{sec:model}
Let $V$ be a collection of robots deployed in a planar environment.  
The \emph{pose} (aka kinematic state) of robot $u \in V$ at time $t \in 
\mathbb{R}^+$ is described by a tuple 
$pose_{u_t}=\tuple{p_{v_t},\phi_{u_t}}$ where $p_{u_t} \in \mathbb{R}^2$ 
represents the \emph{position} of robot $u$ at time $t$, and $\phi_{v_t} 
\in [0,2\pi)$ represents the \emph{orientation} of robot $u$ at time 
$t$.  Robots do \emph{not} know their position or orientation.

Each robot has its own local coordinate system which changes as a 
function of its pose.
Specifically, at time $t$ the local coordinate system of robot $u$ has 
the origin at its own position $p_{u_t}$ and has the $x$-axis aligned 
with its own orientation $\phi_{u_t}$.
All sensing at a robot is recorded in its local
coordinate system.

For $\theta \in [0,2\pi)$ let $R_\theta$ and $\psi(\theta)$ denote 
rotation matrix of $\theta$ and a unit vector of angle $\theta$.
The position of robot $w$ at time $t'$ in the local coordinate system of 
robot $u$ at time $t$ is defined as $p_{w_{t'}} |_{u_t}=R_{-\phi_{u_t}} 
(p_{w_{t'}}-p_{u_t}) = \norm{p_{w_{t'}}-p_{u_t}} 
\psi(\theta_{w_t'}|_{u_t})$, and the orientation of robot $w$ at time 
$t'$ in the local coordinate system of robot $u$ at time $t$ is defined 
as $\phi_{w_{t'}}|_{u_t}=\phi_{w_{t'}}-\phi_{u_t}$.  Hence the pose of 
robot $w$ at time $t'$ in the local coordinate system of robot $u$ at 
time $t$ is described by the tuple $pose_{w_{t'}}|_{u_t} = 
\tuple{p_{w_{t'}}|_{u_t},\phi_{w_{t'}}|_{u_t}}$.

\begin{figure}[tpb]
  \centering
  \def\svgwidth{\linewidth}
  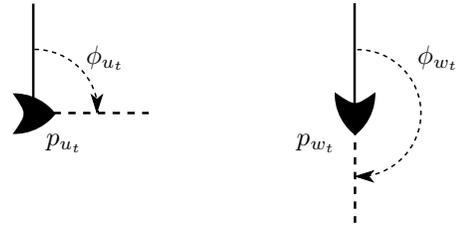
  \caption{In the global coordinate system robot $u$ is pointing right 
    and robot $w$ is pointing down. In robot $u$'s local coordinate 
    system robot $p_{w_t}$ is in front of robot $p_{u_t}$, and in robot 
    $w$'s local coordinate system robot $p_{u_t}$ is to the right of 
    robot $p_{w_t}$.}
  \label{fig:coords}
\end{figure}

The communication graph at time $t$ is a directed graph $G_t=(V,E_t)$, 
where $E_t\subseteq V\times V$ as a set of of directed edges such that 
$(u,v) \in E_t$ if and only if a message sent by robot $u$ at time $t$ 
is received by robot $v$. The neighbors of robot $u$ at time $t$ are the 
set of robots from which $u$ can receive a message at time $t$, denoted 
by $N_{u_t} = \set{v\mid (v,u) \in E_t}$.

For simplicity and ease of exposition, it is assumed that computation, 
communication and sensing proceeds in synchronous lock-step rounds 
$\set{1,2,\ldots}$.  In practice synchronizers~\cite{synchronizers} can 
be used to simulate perfect synchrony in any partially synchronous 
system.
If robot $u$ receives a message from robot $w$ at round $i$ then robot 
$u$ can
identify the message originated from $w$, and estimate the distance 
$\norm{p_{v_i}-p_{w_i}}=d_i{(u,w)}$\footnote{
Many swarm of platforms, including the Kilobots\cite{kilobots}, use the 
same hardware (i.e., infrared transceivers) as a cost-effective way to 
implement both communication and sensing.}.

Robots are capable of using odometry to estimate their pose change 
between rounds in their own local coordinate system. Specifically at 
round $j$ a robot $u \in V$ can estimate its translation change 
$p_{u_i}|_{u_j}$ with respect to round $i<j$ and its orientation change 
$\phi_{u_i}|_{u_j}$ with respect to round $i<j$.
It is assumed that odometry estimates are reliable over intervals of two 
or three rounds (i.e. $i>=j-3$), but suffer from drift over longer time 
intervals.

\subsection{Problem Formulation}
\label{sec:problem}

Formally, the problem statement requires that at every round $i$, each 
robot $u$ computes the relative pose $pose_{w_i}|_{u_i}$ of every 
neighboring robot $w \in N_{u_i}$.
Robots can only perceive each other through distance sensing. For a 
robot $u$ to compute the pose of a neighboring robot $w$ at a particular 
round, it must rely on the distance measurements and communication graph 
in the previous rounds, as well as the odometry estimates of $u$ and $w$ 
in previous rounds.

The algorithms considered do \emph{not} require controlling the motion 
performed by each robot, which allows these algorithms to be run 
concurrently with any motion control algorithm.
Moreover, the algorithms are tailored for large swarms of simple robots,
and as such the size of the messages or the computation requirements do 
not depend on global parameters such as the size or diameter of the 
network.

\section{Localization without Coordination}
\label{sec:alg1}

This section describes a distributed localization algorithm that 
requires no motion coordination between robots and uses minimal 
communication. Each robot localizes its neighbors by finding the 
solution to a system of non-linear equations. For simplicity, this 
section assumes that distance sensing and odometry estimation is perfect 
(e.g. noiseless). Section~\ref{sec:exp} describes how the results 
presented here can be easily extended to handle noisy measurements.

Consider any pair of robots $a$ and $b$ for a contiguous interval of 
rounds $I \subset \mathbb{N}$.
To simplify notation let $p_{a_j \to b_k} = p_{b_k}-p_{a_j}$ denote the 
vector, in the global coordinate system, that starts at $p_{a_j}$ and 
ends at $p_{b_k}$.


\begin{figure}[htpb]
  \centering
  \def\svgwidth{\linewidth}
  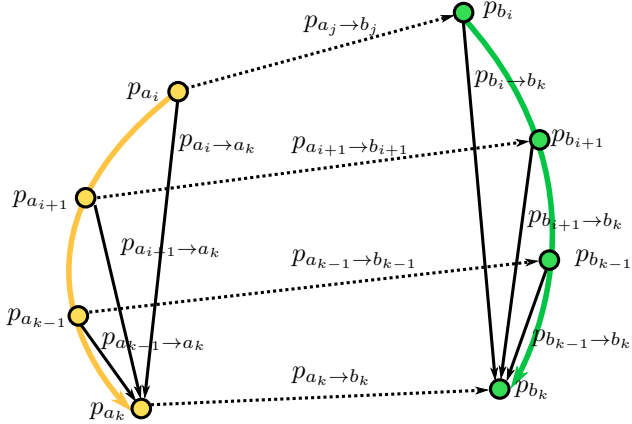
  \caption{Robot $a$ and $b$ in rounds $I=\set{i,\ldots,k}$.}
  \label{fig:tworobots}
\end{figure}

Observing Fig.~\ref{fig:tworobots} it is easy to see that starting at 
$p_{a_i}$ (and in general starting at any $p_{a_j}$ for some $j < k$) 
there are at least two ways to arrive to $p_{b_k}$. For instance, by 
first traversing a dotted line and then a solid line or vice versa.  
Indeed, this holds since by definition for all $j \le k$ we have:

\begin{align}
  p_{a_j \to a_k} + p_{a_k \to b_k} = p_{a_j \to b_k} = p_{a_j \to b_j} + p_{b_j \to b_k}.
  \label{eq:cycle}
\end{align}

For $j=k$ the equation~\ref{eq:cycle} is vacuously true, and for $j < k$ 
this equation can be massaged to express a constraint on the relative 
pose of robots $a$ and $b$ in terms of quantities that individual robots 
can either sense or compute.

\begin{align}
  p_{a_j \to a_k} - p_{b_j \to b_k} + p_{a_k\to b_k} &= p_{a_j \to b_j}\nonumber \\
  -R_{\phi_{a_k}} p_{a_j}|_{a_k} + R_{\phi_{b_k}} p_{b_j}|_{b_k} + 
  R_{\phi_{a_k}} p_{b_k}|_{a_k} &= R_{\phi_{a_j}} p_{b_j}|_{a_j} \nonumber \\
  p_{a_j}|_{a_k} + R_{\phi_{b_k}|_{a_k}}  p_{b_j}|_{b_k} + 
  p_{b_k}|_{a_k} &= R_{\phi_{a_j}-\phi_{a_k}} p_{b_j}|_{a_j}  \nonumber 
  \\
  \norm{p_{a_j}|_{a_k} + R_{\phi_{b_k}|_{a_k}}  p_{b_j}|_{b_k} + 
    p_{b_k}|_{a_k}} &= \norm {p_{b_j}|_{a_j}} \nonumber \\
  \norm{-p_{a_j}|_{a_k} + R_{\phi_{b_k}|_{a_k}}  p_{b_j}|_{b_k} + 
    d_i{(a,b)} \psi(\theta_{b_k}|_{a_k})}
    &= d_j{(a,b)}
  \label{eq:hard}
\end{align}

Dissecting equation~\ref{eq:hard}; $d_j{(a,b)}$ and $d_k{(a,b)}$ are 
known and correspond to the estimated distance between robot $a$ and
$b$ at round $j$ and $k$ respectively; $p_{a_j}|_{a_k}$ and 
$p_{b_j}|_{b_k}$, are also known, and correspond to the odometry 
estimates from round $j$ to round $k$ taken by robot $a$ and $b$ 
respectively; finally $\phi_{b_k}|_{a_k}$ and $\theta_{b_k}|_{a_k}$ are 
both unknown and correspond to the relative position and orientation of 
robot $b$ at round $k$ in the local coordinate system of robot $a$ at 
round $k$.

Considering equation~\ref{eq:hard} over a series of rounds yields a 
non-linear system that, if well-behaved, allows a robot to estimate the 
relative pose of another. The following distributed algorithm leverages 
the constraints captured by this system to allow every robot to compute 
the relative pose of its neighbors.

\begin{algorithm}[htpb]
  \begin{algorithmic}[1]
    \For{{\bf each} robot $u \in V$ and every round $k \in 
      \set{1,\ldots}$}
      \State {\bf broadcast} $\tuple{p_{u_{k-1}}|_{u_k}, 
        \phi_{u_{k-1}}|_{u_k}}$
      \State {\bf receive} $\tuple{p_{w_{k-1}}|_{w_k}, 
        \phi_{w_{k-1}}|_{w_k}}$ for $w \in N_{u_k}$
      \State $I = \set{k-\delta,k}$
      \For{{\bf each} $w \in \bigcap_{j \in I} N_{u_j}$}
        \State integrate odometry $p_{u_j}|_{u_k}, \phi_{u_j}|_{u_k}$ 
        for $j \in I$
        \State find $\hat{\theta}_{w_k}|_{u_k},\hat{\phi}_{w_k}|_{u_k}$ 
        such that (\ref{eq:hard}) holds $\forall j \in I$
        \State $pose_{w_k}|_{u_k} \gets 
        \tuple{d_k{(u,w)}\psi(\hat{\theta}_{w_k}|_{u_k}),\hat{\phi}_{w_k}|_{u_k}}$
      \EndFor
    \EndFor
  \end{algorithmic}
  \caption{Localization without Coordination}
\end{algorithm}

At each round of Algorithm~1 every robot sends a constant amount of 
information (its odometry measurements for that round) and therefore its 
message complexity is $O(1)$. The computational complexity of 
Algorithm~1 is dominated by solving the system of non-linear equations 
(line 7), which can be done by numerical methods~\cite{lm2010} in 
$O(\varepsilon^{-2})$ where $\varepsilon$ is the desired accuracy.

The parameter $\delta$ in (line 4) of Algorithm~1 corresponds to the 
number of rounds over which equation~\ref{eq:hard} is considered.  Since 
there are two unknowns then to avoid an undetermined system it must be 
required that $\delta \ge 2$, and it will be shown that in practice 
$\delta=2$ suffices.

\begin{figure}[htpb]
  \begin{subfigure}[b]{0.5\linewidth}
    \centering
    \def\svgwidth{\linewidth}
    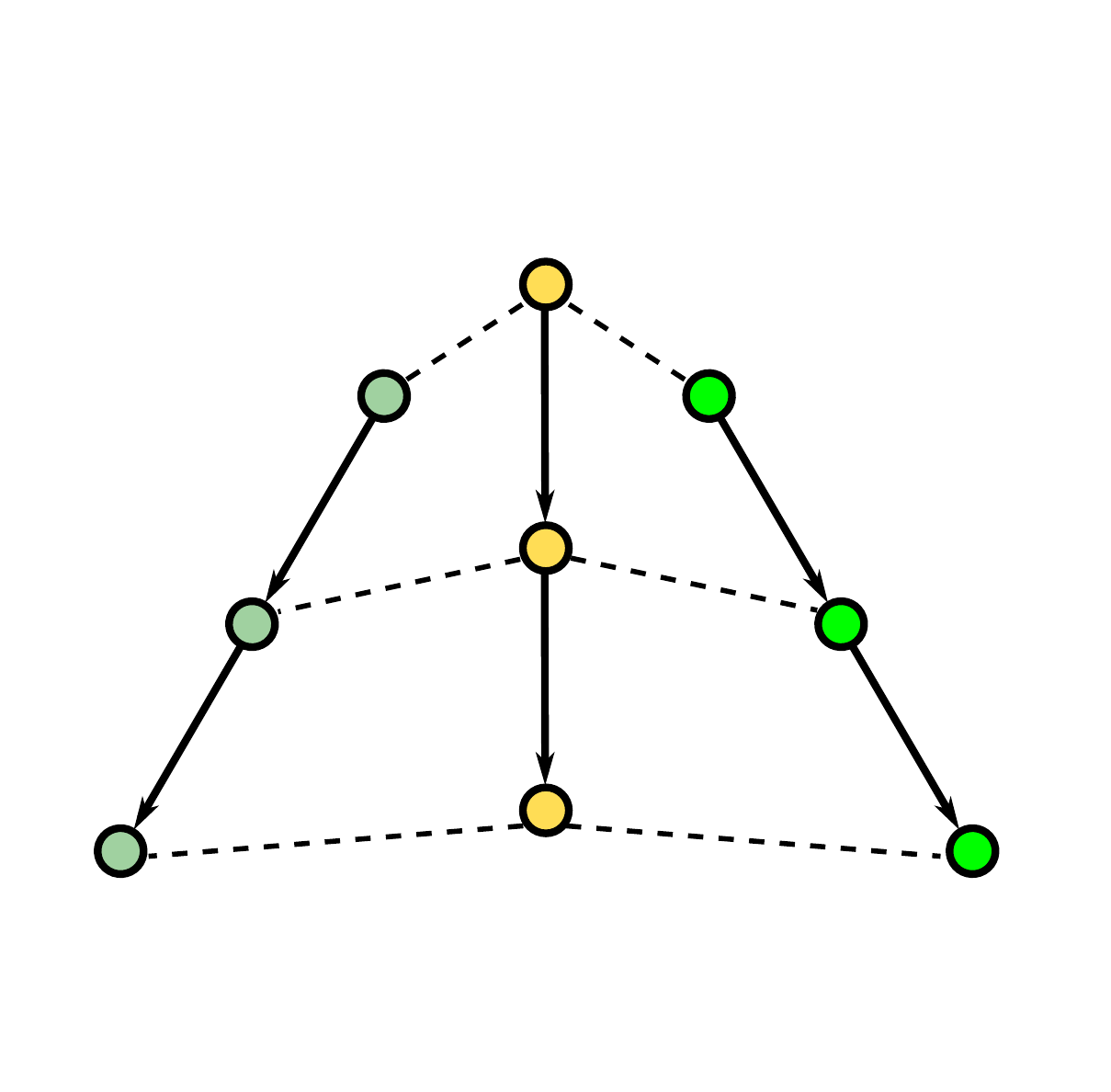
    \caption{Flip Ambiguity}
    \label{fig:degenerate1}
  \end{subfigure}%
  \begin{subfigure}[b]{0.5\linewidth}
    \centering
    \def\svgwidth{\linewidth}
    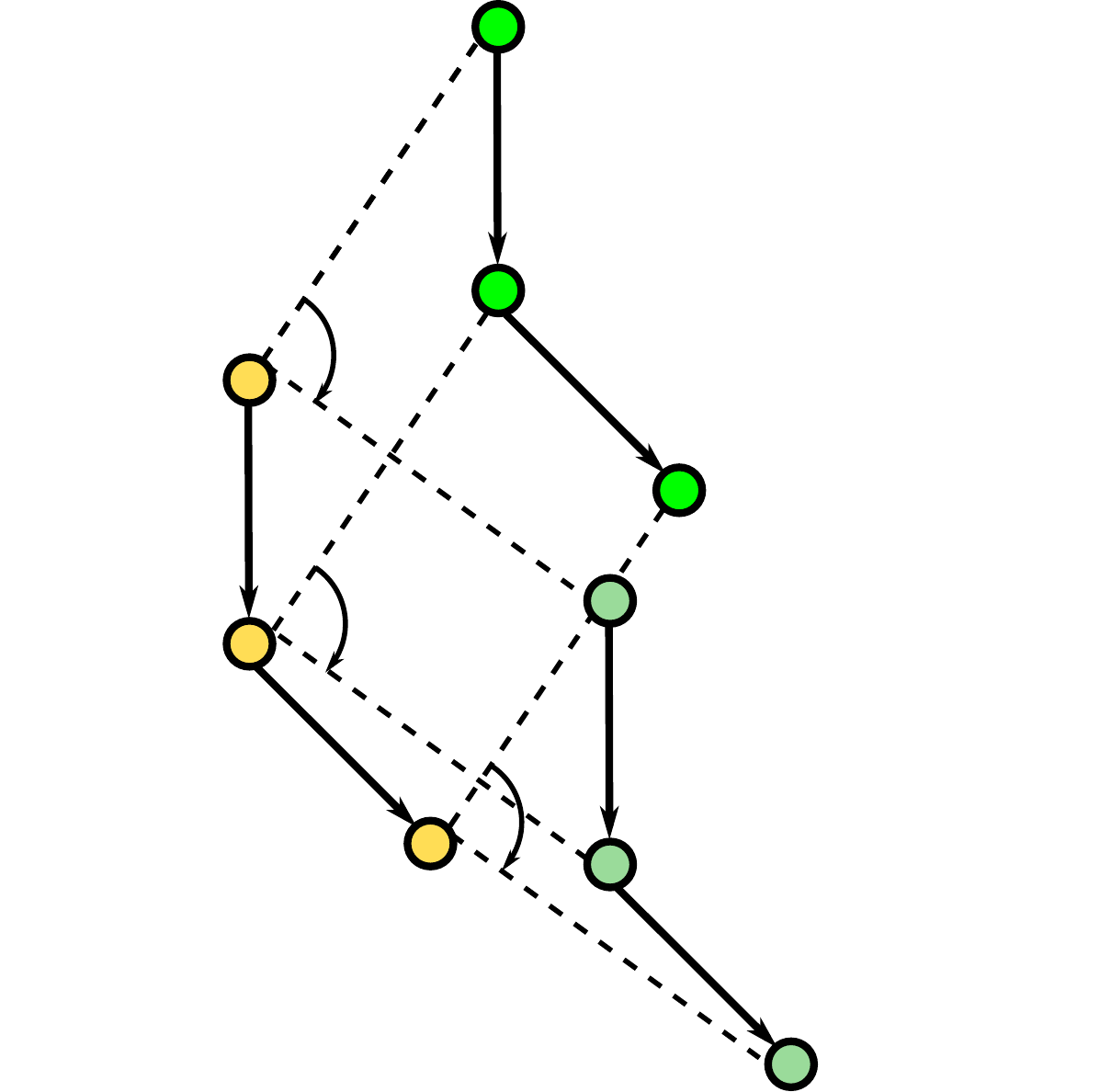
    \caption{Rotation Ambiguity}
    \label{fig:degenerate2}
  \end{subfigure}
  \caption{Yellow robot cannot fully resolve the relative position of 
    green robot using the available distance measurements due to 
    degenerate motions.  }
  \label{fig:degenerate}
\end{figure}

Regardless of the choice of $\delta$ there are motion patterns for which 
any algorithm that does not enforce a very strict motion coordination 
(which includes Algorithm~1, which enforces no motion coordination) 
cannot recover the relative pose of neighboring robots.  These motions 
are referred to as \emph{degenerate}, and are described next (see 
Fig.~\ref{fig:degenerate}).
First, if during $\delta$ rounds two robots follow a linear trajectory, 
then the relative pose between these robots can only be recovered up to 
a \emph{flip ambiguity}. 
%
Second, if during $\delta$ rounds one robot follows a displaced version 
of the trajectory followed by another robot, then it is possible to 
infer the relative orientation of the robots, but a \emph{rotation 
ambiguity} prevents the recovery of the relative position. A degenerate 
motion can be a flip ambiguity, a rotation ambiguity, or a combination 
of both.

Fortunately degenerate motions are rare. More precisely degenerate 
motions are a set of measure zero (for example, this implies that if the 
motions are random, then with probability 1 they are not degenerate).  
This can be shown to be a consequence of the generic rigidity of a 
triangular prism in Euclidean 2-space, see~\cite{rigidity99} for a 
thorough treatment of rigidity.
We conclude this section with the following theorem, which formalizes 
the properties of Algorithm~1.

\begin{thm}
  If at round $i$, robots $u$ and $w$ have been neighbors for a 
  contiguous interval of $\delta$ or more rounds, and perform 
  non-degenerate motions, then at round $i$ Algorithm~1 computes 
  $pose_{w_i}|_{u_i}$ at $u$ and $pose_{u_i}|_{w_i}$ at $w$.
\end{thm}

%
%
%
%
%

\section{Localization with Coordination}
\label{sec:alg2}

This section describes a distributed localization algorithm that uses a 
simple stop/move motion coordinate scheme, and requires communication 
proportional to the number of neighbors. Using the aforementioned motion 
coordination scheme allows robots to compute the relative pose of
neighboring robots through trilateration with no sensing errors.
Section~\ref{sec:exp} generalizes this to consider noise.

By collecting multiple distance estimates a moving robot can use 
trilateration to compute the relative position of a stationary robot.  
Two such distance estimates already suffice to allow the moving robot to 
compute the relative position of a stationary robot up to a flip 
ambiguity (i.e., a reflection along the line that passes through the 
coordinates at which the measurements were taken).

\begin{figure}[htpb]
  \centering
  \def\svgwidth{\linewidth}
  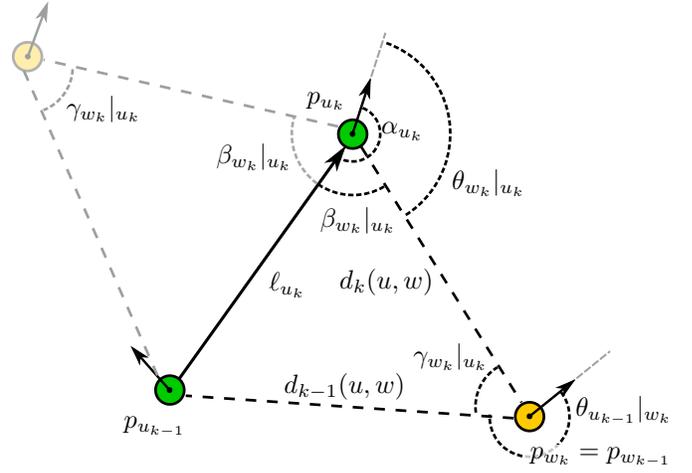
  \caption{Moving robot (green) uses trilateration to compute the 
    relative position of stationary robot (yellow) up to a flip 
    ambiguity.}
  \label{fig:trilateration}
\end{figure}

Consider two neighboring robots $u$ and $w$ where from round $k-1$ to 
round $k$ robot $u$ moves while robot $w$ remains stationary (see 
Fig.~\ref{fig:trilateration}). Robot $u$ can compute the relative 
position $p_{w_k}|_{u_k}$ of robot $w$ at round $k$ up to a flip 
ambiguity, relying only on the distance measurements to robot $w$ at 
round $k-1$ and round $k$, and its own odometry for round $k$.  
Specifically the cosine yields the following.

\begin{align}
   \ell_{u_k} &= \norm{p_{u_{k-1}}|_{u_k}} \quad\quad \alpha_{u_k} = 
   \measuredangle (p_{u_{k-1}}|_{u_k}) \nonumber \\
\beta_{w_k}|_{u_k} &= \cos^{-1}\paren{\frac{\ell_{u_k}^2 + d^2_{k}(u,w) 
    - d^2_{k-1}(u,w)}{2\ell_{u_k} d_k(u,w)}} \\
\gamma_{w_k}|_{u_k} &= 
\cos^{-1}\paren{\frac{d^2_k(u,w)+d^2_{k-1}(u,w)-\ell^2_{u_k}}{2 d_k(u,w) 
    d_{k-1}(u,w)}} \\
\theta_{w_k}|_{u_k} &= \alpha_{u_k} \mp \beta_{w_k}|_{u_k} \\
\theta_{u_k}|_{w_k} &= \theta_{u_{k-1}}|_{w_k} \pm \gamma_{w_k}|_{u_k}
\label{eq:cor}
\end{align}

In order for robot $u$ to fully determine the relative pose of robot $w$ 
at round $k$ (ignoring the flip ambiguity) it remains only to compute 
$\phi_{w_k}|_{u_k}$. Observe that given knowledge of 
$\theta_{u_{k-1}}|_{w_k}$, robot $u$ can leverage Eq.~\ref{eq:cor} to 
compute $\theta_{u_k}|_{w_k}$ using the correction term 
$\gamma_{w_k}|_{u_k}$ computed through the cosine law. The following 
identity can be leveraged to easily recover $\phi_{w_k}|_{u_k}$ using 
$\theta_{u_k}|_{w_k}$ and $\theta_{w_k}|_{u_k}$.

\begin{align}
  \phi_{u_k}|_{w_k} = \theta_{w_k}|_{u_k} - \theta_{u_k}|_{w_k} + \pi \imod{2\pi}
\end{align}

Summing up, if robot $u$ moves from round $k-1$ to round $k$ while robot 
$w$ remains stationary, then using $d_{k-1}(u,w)$, $d_k(u,w)$ and 
$p_{u_{k-1}}|_{u_k}$ robot $u$ can compute the relative position of
robot $w$ at time $k$. Additionally, if knowledge of 
$\theta_{u_{k-1}}|_{w_k}$ is available robot $u$ can also compute the 
relative orientation of robot $w$ at time $k$. Both the position and 
orientation are correct up to a flip ambiguity.

A robot can resolve the flip ambiguity in position and orientation by 
repeating the above procedure and checking for consistency of the 
predicted position and orientation. We refer to motions which preserve 
symmetry and therefore prevent the flip ambiguity from being resolved 
(for instance, collinear motions) as degenerate.

\begin{figure}[htpb]
  \begin{subfigure}[b]{0.5\linewidth}
    \centering
    \def\svgwidth{\linewidth}
    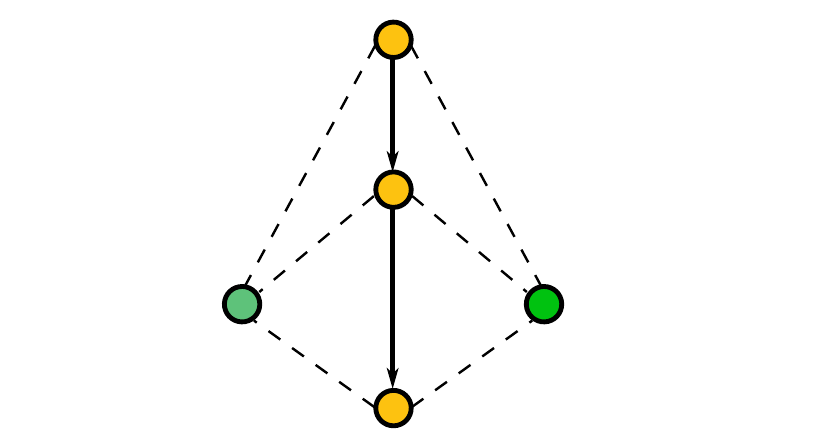
    \caption{Flip Ambiguity}
    \label{fig:triangulate1}
  \end{subfigure}%
  \begin{subfigure}[b]{0.5\linewidth}
    \centering
    \def\svgwidth{\linewidth}
    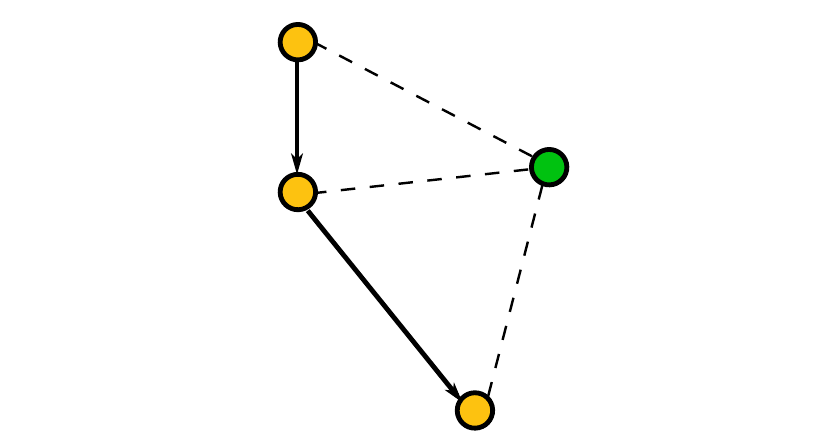
    \caption{Unambiguous}
    \label{fig:triangulate2}
  \end{subfigure}
  \caption{Moving robot (yellow) localizing a stationary robot (green) 
    using distance measurements (dashed lines) and odometry (solid 
    arrows).}
  \label{fig:triangulate}
\end{figure}

To bootstrap the previous trilateration procedure and allow robot $u$ to 
recover the orientation of robot $w$, robot $w$ ---which remains 
stationary from round $k-1$ to round $k$--- must somehow compute 
$\theta_{u_{k-1}}|_{w_{k-1}} = \theta_{u_{k-1}}|_{w_{k}}$ and 
communicate it to robot $u$ by round $k$.


Note that the distance measurements between a stationary robot and a 
moving robot are invariant to rotations of the moving robot around the 
stationary robot. This prevents a stationary robot from recovering the 
relative position of a moving neighbor using any number of distance 
estimates. 

\begin{figure}[htpb]
  \centering
  \includegraphics[width=\linewidth]{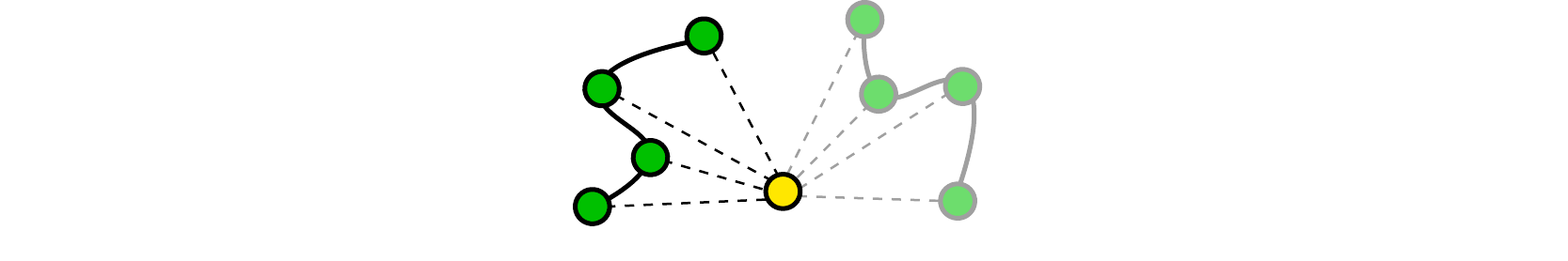}
  \caption{Stationary robot (yellow) cannot compute the relative 
    position of the moving robot (green), since all distance 
    measurements (dashed lines) are invariant to rotations around the 
    stationary robot.}
\end{figure}

Therefore to successfully use the aforementioned trilateration procedure 
requires coordinating the motion of the robots
in a manner that gives every robot a chance to move and ensures that 
when a robot is moving its neighbors remain stationary.  Formally, a 
\emph{motion-schedule} is an algorithm that at each round classifies 
every robots as being either mobile or stationary. A motion-schedule is 
\emph{well-formed} if at every round $i$ the set of robots classified as 
mobile define an independent set of the communication graph $G_i$ (i.e.  
no two mobile robots are neighbors).
The \emph{length} of a motion-schedule is the maximum number of rounds 
that any robot must wait before it is classified as mobile. A 
motion-schedule is \emph{valid} if it is well-formed and has finite 
length.

The validity of a motion-schedule ensures that mobile robots can use
trilateration to compute the relative positions of all its neighbors, 
and having a motion-schedule of finite length guarantees every robot 
gets a chance to move. The next subsection provides a description of a 
distributed algorithm that produces a valid motion-schedule.
Algorithm~2 describes a distributed localization algorithm that 
leverages a valid motion-schedule and trilateration.

\begin{algorithm}[htpb]
  \begin{algorithmic}[1]
    \State $\Theta_{u_0} \gets \emptyset$ $\forall u \in V$
    \For{{\bf each} robot $u \in V$ and every round $k \in 
      \set{1,\ldots}$}
      \State {\bf broadcast} $\tuple{p_{u_{k-1}}|_{u_k}, 
        \phi_{u_{k-1}}|_{u_k}, \Theta_{u_{k-1}}}$
      \State {\bf receive} $\tuple{p_{w_{k-1}}|_{w_k}, 
        \phi_{w_{k-1}}|_{w_k}, \Theta_{u_{k-1}}}$ for $w \in N_{u_k}$
      \If{state = mobile}
        \State $\Theta_{u_{k}} \gets \set{\hat{\theta}_{w_k}|_{u_k} 
          \mbox{ through Eq.~(4-5)}}$
        \State $\hat{\phi}_{w_k}|_{u_k} \gets$ use Eq.~(6-7) $\forall w 
        \in N_{u_k}$
        \State use previous state resolve \emph{flip} in $\Theta_{u_{k}}$
      \Else
        \State update $\Theta_{u_k}$ through $\phi_{w_k{-1}}|_{w_k},
        p_{w_{k-1}}|_{w_k}$
      \EndIf
      \State {\small $\hat{pose}_{w_k}|_{u_k} \gets 
        \tuple{d_k{(u,w)}\psi(\hat{\theta}_{w_k}|_{u_k}),\hat{\phi}_{w_k}|_{u_k}}$
       $\forall w \in N_{u_k}$}
      \State state $\gets$ \textsc{motion-scheduler}
      \If{state = mobile}
      \State move according to \textsc{motion-controller}
      \Else
      \State remain stationary
      \EndIf
    \EndFor
  \end{algorithmic}
  \caption{Localization with Coordination}
\end{algorithm}

At each round of Algorithm~2 every robot sends a message containing its 
own odometry estimates and $\Theta_{u_{k-1}}$, which is the set of 
previous position estimates (one for each of its neighbor), and 
therefore its message complexity is $O(\Delta)$.
Mobile robots use trilateration to compute the relative position and 
relative orientation of its neighbors, and when possible stationary 
robots update the relative position and orientation of mobile robots 
using the received odometry estimates. In either case, the amount of 
computation spent by Algorithm~2 to localize each robot is constant.

\begin{thm}
  (Assuming a valid motion-schedule.)
  If at round $i$, robots $u$ and $w$ have been neighbors for a 
  contiguous set of rounds during which robot $u$ performed a 
  non-degenerate motion, then at round $i$ Algorithm~2 computes 
  $pose_{w_i}|_{u_i}$ at $u$.
\end{thm}

\subsection{Motion Scheduling}

As a straw-man distributed algorithm that requires no communication and 
outputs a valid motion-schedule, consider an algorithm that assigns a 
single mobile robot to each round, in a round robin fashion (i.e.  at 
round $i$ let robot $k=i\mod n$ be mobile and let the remaining $n-1$ 
robots be stationary). Although the motion-schedule produced by such an 
algorithm is valid, it is not suited for a swarm setting, since it 
exhibits no parallelism and the time required for a robot to move is 
linear on the number of robots.

Finding a motion-schedule that maximizes the number of mobile robots at 
any particular round is tantamount to finding a maximum independent set 
(aka MaxIS) of the communication graph, which is NP-hard. Similarly, 
finding a motion-schedule with minimal length implies finding a 
vertex-coloring with fewest colors of the communication graph, which is 
also NP-hard.

Algorithm~3 describes a motion-schedule with the more modest property of 
having the set of moving robots at each round define a maximal 
independent set (aka MIS) of the communication graph. Once a robot is 
classified as being mobile, it does not participate on subsequent MIS 
computations, until each of its neighbors has also been classified as 
mobile. Given these properties, it is not hard to show that for any 
robot $u$ and a round $k$, the number of rounds until robot $u$ is 
classified as mobile is bounded by the number of neighbors of robot $u$ 
at round $k$.

\begin{algorithm}[htpb]
  \begin{algorithmic}[1]
    \If{$\forall w \in N_u$ state$_w =$ inactive}
    \State state$_u \gets$ compete
    \EndIf
    \If{state$_u = $ compete}
    \If{$u$ is selected in distributed MIS}
      \State state$_u \gets$ inactive
      \State {\bf output} mobile
    \EndIf
    \EndIf
    \State {\bf output} stationary
  \end{algorithmic}
  \caption{Motion-Scheduler}
\end{algorithm}


\begin{thm}
  Algorithm~3 defines a valid motion-schedule with length $\Delta+1$.
\end{thm}

The description of Algorithm~3 utilizes a distributed MIS algorithm as a 
subroutine (line 4 in the pseudo-code). However, it should be noted that 
the problem of finding an MIS with a distributed algorithm is a 
fundamental symmetry breaking problem and is far from trivial.  
Fortunately, the MIS problem has been studied extensively by the 
distributed computing community, and extremely efficient solutions have 
been proposed under a variety of communication 
models~\cite{luby86,podc08,disc11}. The classic solution~\cite{luby86} 
requires $O(\log n)$ communication rounds
and every node uses a total of $O(\log n)$~\cite{sirocco09} bits of 
communication. For a wireless network settings, it is 
known~\cite{podc08} how to find an MIS exchanging at most $O(\log^* 
n)$\footnote{The iterated logarithm function counts the number of times 
  the logarithm is applied to the argument before the result is less or 
  equal to 1. It is an extremely slowly growing function, for instance 
  the iterated logarithm of the number of atoms in the universe is less 
  than 5.} bits.
Due to lack of space, for the purposes of this paper it should suffice 
to know that it is possible to implement a distributed MIS protocol in 
the lower communication layers without significant overhead.

\section{Algorithm Evaluation}
\label{sec:exp}

To evaluate the performance of the proposed localization algorithms, 
this section considers a generalization of the system model described in 
Section~\ref{sec:model} where the distance estimates and the odometry 
estimates are subject to noise from an independent zero-mean 
distribution. In particular we assume multi-variate Gaussian noise with 
zero-mean and covariance matrix $\Sigma = diag(\sigma_d, \sigma_\phi, 
\sigma_x, \sigma_y)$.

Algorithm~1 relied on finding a zero in a non-linear system of equations 
constructed using the distance estimates and odometry estimates 
pertinent to that robot.
When these estimates are subject to noise, the corresponding non-linear 
system is no longer guaranteed to have a zero. To cope with noisy 
measurements it suffices to instead look for the point that minimizes 
the mean-squared error. This incurs in no additional computational 
overhead, since it can be accomplished using the same numerical methods 
used in the noiseless case.

To understand the sensitivity of Algorithm~1 to the various sources of 
error (distance estimates, orientation odometry and translation 
odometry), we carried out extensive simulations.
For each simulation trial robots are deployed randomly in a region of 
10m x 10m, and at each round each robot is allowed to perform a motion 
with a random orientation change between $[-\pi/4,\pi/4]$ and a 
translation change which is normally distributed with a mean of 3m and a 
variance of 0.5m.
The length of each trial is 20 rounds. The plots below show the mean 
squared error (MSE) in the computed position (blue) and orientation 
(red) over 50 random trials for various different noise parameters.  
Since to initialize the position and orientation estimates Algorithm~2 
requires at least three rounds, the first three rounds of every trial 
were discarded.

\begin{figure}[htpb]
  \centering
  \includegraphics[width=\linewidth]{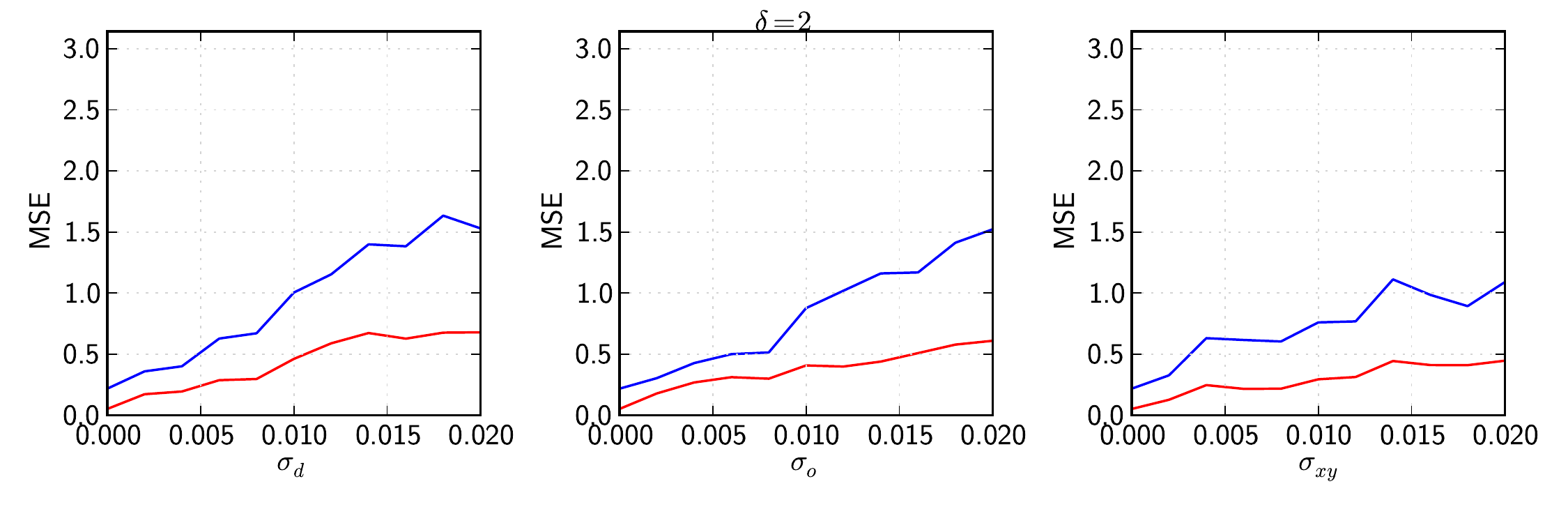}
  \includegraphics[width=\linewidth]{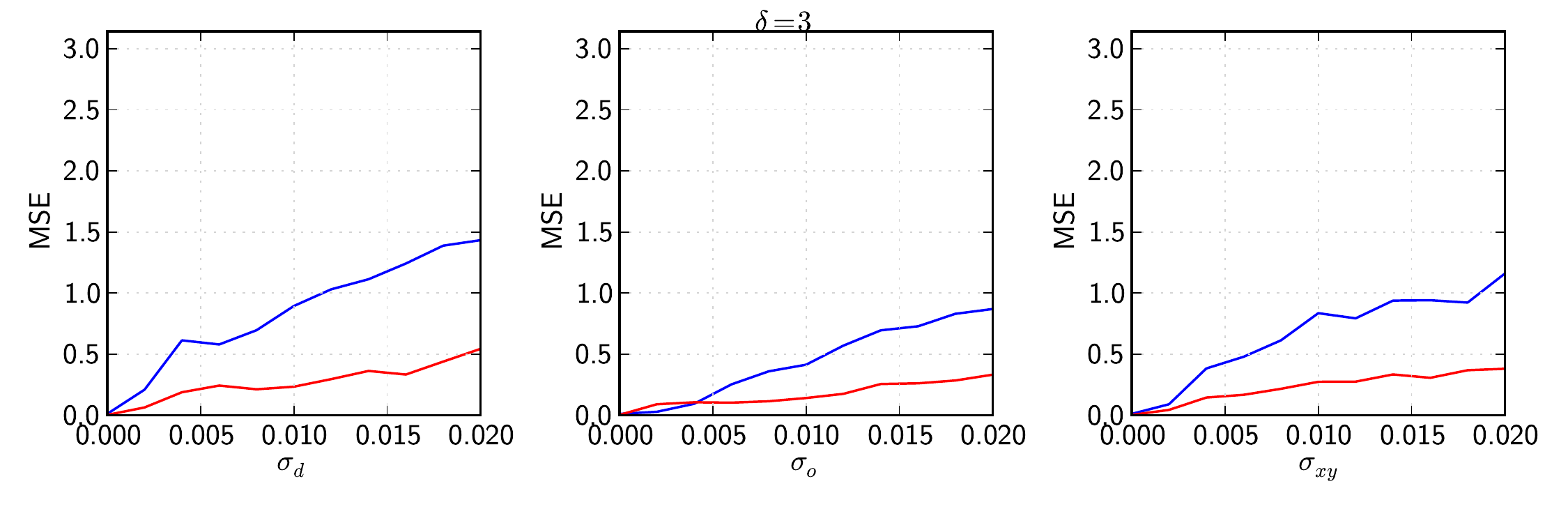}
  \caption{Each plot shows MSE of the position (blue) and orientation 
    (red) as a function of one component of the variance $\Sigma$.
    From left to right, each column shows the MSE as a function of 
    $\sigma_d$, $\sigma_o$ and $\sigma_{x,y}$. The top row shows the 
    results with $\delta=2$ and the bottom row for $\delta=3$. }
\end{figure}

Not surprisingly the results produced by Algorithm~1 are sensitive to 
errors in all axis, although it is slightly more robust to errors in the 
translation odometry than in the distance sensing.
Furthermore, the relative orientation estimate was consistently more 
tolerant to noise than the position estimate.
As it would be expected, for all the different noise settings, 
increasing the parameter $\delta$ from $2$ to $3$ consistently reduced 
the MSE in both position and orientation produced by Algorithm~1.  
However, increasing $\delta$ also increases the computational costs of 
the algorithm and only gives diminishing returns.

In the case of Algorithm~2, to perform trilateration using estimates 
subject to zero-mean noise corresponds we instead perform trilateration 
using the expected value of the estimates conditioned on the information 
available.

To understand the sensitivity of Algorithm~2 to the different sources of
error, we used the same simulation environment and parameters as with 
Algorithm~1, with one exception.
Namely, to keep the number of motions per trial for Algorithm~1 and 
Algorithm~2 roughly the same, the length of the trial was doubled, since 
at each round, for every pair of nodes, only one of them will be mobile 
and the other will remain stationary.

\begin{figure}[htpb]
  \centering
  \includegraphics[width=\linewidth]{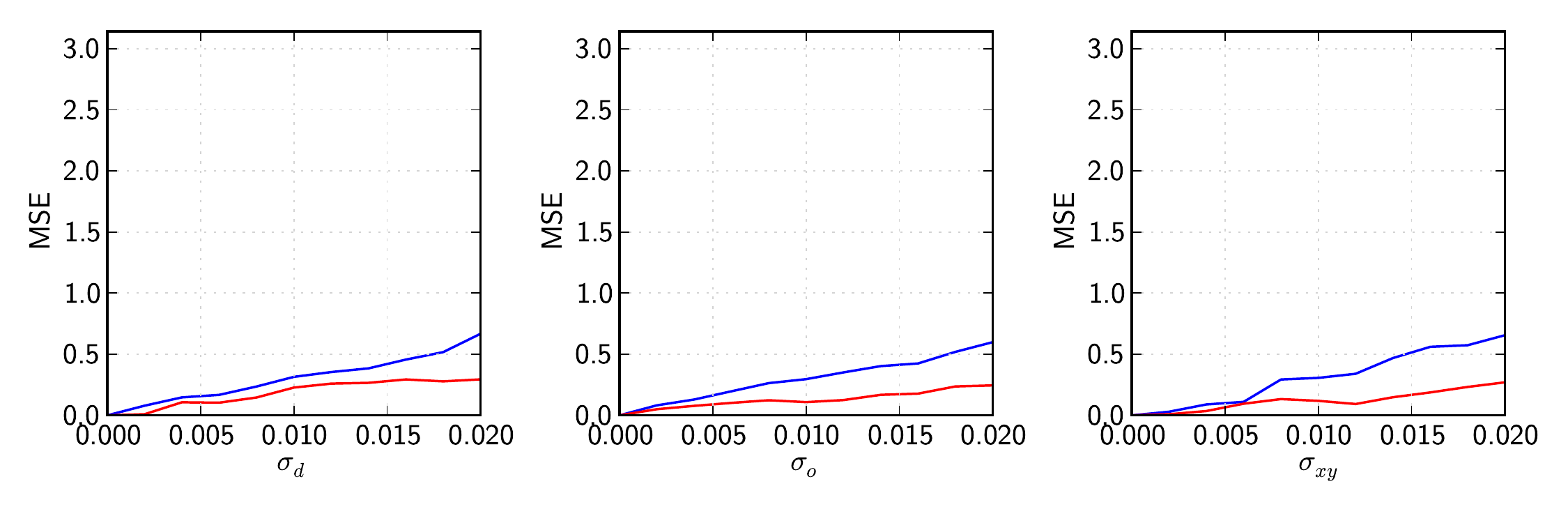}
  \caption{Each plot shows MSE of the position (blue) and orientation 
    (red) as a function of one component of the variance $\Sigma$.
    From left to right, each column shows the MSE as a function of 
    $\sigma_d$, $\sigma_o$ and $\sigma_{x,y}$.}
\end{figure}

The pose estimates produced by Algorithm~2 are for the most part equally 
affected by noise in either of the dimension. As it was the case with 
Algorithm~1, the relative orientation estimate was consistently more 
tolerant to noise than the position estimate.  Overall compared to 
Algorithm~1, the results show that Algorithm~2 is in all respects less 
sensitive to noise.

\subsection{Motion Control and Localization}

To conclude we empirically explore the feasibility of composing existing 
motion control algorithms with the proposed localization algorithms. For 
its simplicity we consider the canonical problem of 
flocking~\cite{flock}.
Informally, flocking describes an emergent behavior of a collection of 
agents with no central coordination that move cohesively despite having 
no common a priori sense of direction.

Flocking behavior has received a lot of attention in the scientific 
community.
Vicsek et al.~\cite{vicsek} studied flocking
from a physics perspective through simulations. The work of Vicsek et 
al. focused on the emergence of alignment in self-driven particle 
systems. Flocking has also been studied from a control theoretic 
perspective, for example in the work of Olfati-Saber~\cite{olfati} and 
Jadbabaie et al.~\cite{jadbabaie}, where the emphasis is on the 
robustness of the eventual alignment process despite the local and 
unpredictable nature of the communication.

For the purposes of this section we consider the standard and most 
simplistic flocking behavior, where each robot aligns its heading with 
its neighbors. Namely, at each round every robot steers its own 
orientation to the average orientation of its neighbors. It has been 
shown~\cite{olfati,jadbabaie} that under very mild assumptions of the 
connectivity of the communication graph, the following procedure 
converges to a state where all robots share the same orientation.

\begin{figure*}[htpb]
  \centering
  \includegraphics[width=\linewidth]{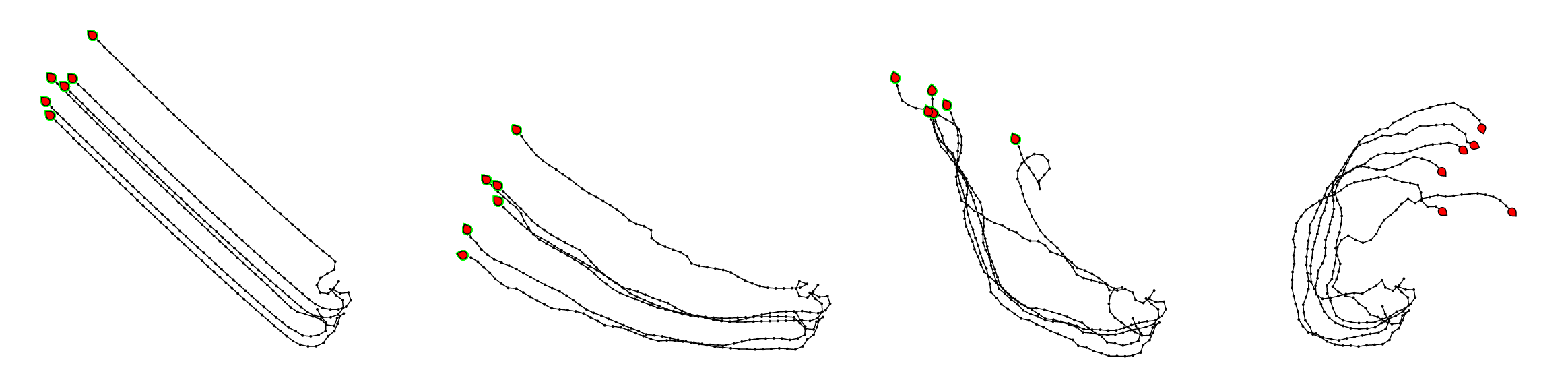}
  \caption{Final configuration of 6 robots after four 40 round runs of a 
    flocking algorithm composed with Algorithm~1 to provide relative 
    position and relative orientation estimates. All runs have the same 
    initial random configuration. From left to right the variance of all 
    noise parameters was increased.}
  \label{fig:flock}
\end{figure*}

Fig.~\ref{fig:flock} (on the following page) shows the results of the 
described average-based flocking algorithm when combined with 
Algorithm~1 to provide relative orientation estimates. Initially the 
first rounds the robots move erratically while the position and 
orientation estimates are initialized, and soon after the orientations 
of all the robots converge. Increasing the error in the distance sensing 
and odometry measurements is translated in greater inaccuracy in the 
resulting relative orientation estimates, which affects the resulting 
flocking state.

\section{Conclusions}

We presented two distributed algorithms to solve the relative 
localization problem tailored for swarms of simple robots. The 
algorithms have different communication and computational requirements, 
as well as different robustness to sensing errors. Specifically, having 
greater communication and coordination allows us to reduce the required 
computational complexity and increase the robustness to sensing errors.
In future work, we hope to further whether this trade-off is inherent to 
the problem or not.

\bibliographystyle{plainnat}
\bibliography{refs}

\end{document}